\documentclass{article}

\usepackage{arxiv}

\usepackage[utf8]{inputenc} 
\usepackage[T1]{fontenc}    
\usepackage{hyperref}       
\usepackage{url}            
\usepackage{booktabs}       
\usepackage{amsfonts}       
\usepackage{nicefrac}       
\usepackage{microtype}      
\usepackage{mathtools}
\usepackage{cleveref}       
\usepackage{lipsum}         
\usepackage{graphicx}
\usepackage[font=small]{subfig}
\usepackage{natbib}
\usepackage{doi}

\usepackage{booktabs}
\usepackage{siunitx}

\title{Unsupervised Learning for Topological Classification of Transportation Networks}

\author{ \href{https://orcid.org/0000-0001-8368-2307}{\includegraphics[scale=0.06]{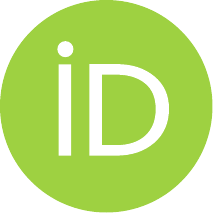}\hspace{1mm}Sina Sabzekar} \\
	Department of Civil Engineering\\
	Sharif University of Technology\\
	Tehran, Iran\\
	\texttt{sina.sabzekar@sharif.edu} \\
	\And
	\href{https://orcid.org/0009-0009-8657-7452}{\includegraphics[scale=0.06]{graphics/orcid.pdf}\hspace{1mm}Mohammad Reza Valipour Malakshah} \\
	Department of Civil Engineering\\
	Sharif University of Technology\\
	Tehran, Iran \\
	\texttt{mohammad.valipoor98@sharif.edu} \\
	\And
	\href{https://orcid.org/0000-0002-0440-6660}{\includegraphics[scale=0.06]{graphics/orcid.pdf}\hspace{1mm}Zahra Amini}\thanks{Corresponding Author} \\ \\
	Department of Civil Engineering\\
	Sharif University of Technology\\
	Tehran, Iran \\
	\texttt{zahra.amini@sharif.edu} \\
}

\renewcommand{\headeright}{ }
\renewcommand{\shorttitle}{Unsupervised Learning for Topological Classification of Transportation Networks}

\hypersetup{
pdftitle={A template for the arxiv style},
pdfsubject={q-bio.NC, q-bio.QM},
pdfauthor={David S.~Hippocampus, Elias D.~Striatum},
pdfkeywords={First keyword, Second keyword, More},
}

\begin{document}
\maketitle

\begin{abstract}
	With increasing urbanization, transportation plays an increasingly critical role in city development. The number of studies on modeling, optimization, simulation, and data analysis of transportation systems is on the rise. Many of these studies utilize transportation test networks to represent real-world transportation systems in urban areas, examining the efficacy of their proposed approaches. Each of these networks exhibits unique characteristics in their topology, making their applications distinct for various study objectives. Despite their widespread use in research, there is a lack of comprehensive study addressing the classification of these networks based on their topological characteristics. This study aims to fill this gap by employing unsupervised learning methods, particularly clustering. We present a comprehensive framework for evaluating various topological network characteristics. Additionally, we employ two dimensionality reduction techniques, namely Principal Component Analysis (PCA) and Isometric Feature Mapping (ISOMAP), to reduce overlaps of highly correlated features and enhance the interpretability of the subsequent classification results. We then utilize two clustering algorithms, K-means and HDBSCAN, to classify 14 transportation networks. The PCA method, followed by the K-means clustering approach, outperforms other alternatives with a Silhouette score of $0.510$, enabling the classification of transportation networks into five clusters. We also provide a detailed discussion on the resulting classification.
\end{abstract}


\section{Introduction}
In the last century, the rapid urbanization process has drawn significant attention to the crucial role of transportation as a fundamental aspect of urban development~\citep{fancello2014modeling, badhrudeen2022geometric}. Transportation systems play a vital role in facilitating the movement of people and goods within cities, directly impacting the economic prosperity of societies. This has led to substantial efforts from both academic and industrial sectors to enhance the efficiency and safety of transportation systems within urban environments. As cities expand, the complexity of urban road networks has increased due to the growth of roads and streets~\citep{sharifi2019resilient}.

Mathematical models of urban transportation networks have provided valuable insights and solutions to various core issues associated with transportation systems. These include travel demand modeling~\citep{hassanzadeh2023using}, traffic flow prediction~\citep{sabzekar2023spatial}, traffic flow assignment~\citep{rahman2023data}, traffic congestion detection~\citep{lu2019congestion}, and traffic signal control~\citep{amini2018data, amini2018development, amini2016estimating}. With the evolution of research methodologies and the emergence of new challenges in the transportation domain, an increasing number of studies aim to model and address problems concerning transportation networks. Consequently, several transportation networks have been extensively employed by researchers to simulate real-world urban transportation networks, differing significantly in size, topology, and geometry. Among these networks, SiouxFalls, a widely used transportation test network, resembles the road network of SiouxFalls, South Dakota~\citep{Dataset_SiouxFalls}. Given the extensive utilization of SiouxFalls and similar transportation networks in various ongoing research projects within the transportation domain~\citep{jayakrishnan1994evaluation, bar2002origin, bar2003origin, boyce2003validation, bar2006solving, bar2007differences, bar2007non, bar2010traffic, bar2012user, bar2013computational, rey2019branch, yu2020day, rahman2023data, liu2023heterogeneous}, there arises a critical need to classify these networks. For instance, when researchers test a new approach on different transportation networks and claim its superiority, it is essential to determine the distinctiveness of these transportation networks. Questions such as whether the differences lie in the network size or what specific features contribute to their similarities or differences need to be addressed. Moreover, the application of centrality indices is constrained by the substantial computational expenses involved, particularly when dealing with large-scale networks such as urban networks. For instance, in study~\citep{badhrudeen2022geometric}, the authors chose to solely utilize degree centrality, omitting the computation of other centrality indices due to their computational demands. Several studies in the existing literature seek to identify specific correlations between certain centralities and network features. Consequently, it becomes imperative to assist these researchers in selecting the most appropriate networks based on the needs and objectives of their studies.

This paper aims to provide answers to these questions and address the mentioned challenges by employing unsupervised learning to classify transportation networks. In contemporary studies, machine learning has demonstrated remarkable effectiveness across various fields\citep{sadatnya2023machine}. Unsupervised learning, a subset of machine learning, operates with unlabeled data, making it ideal for identifying hidden patterns~\citep{barlow1989unsupervised}. Clustering, a key task in unsupervised learning, involves identifying hidden patterns and relationships within data, ultimately grouping data records with similar characteristics~\citep{diday1976clustering}. Various clustering methods, such as K-means~\citep{macqueen1967some}, mini-batch K-means~\citep{sculley2010web}, DBSCAN~\citep{ester1996density}, HDBSCAN~\citep{campello2015hierarchical}, and hierarchical clustering~\citep{johnson1967hierarchical}, have been extensively utilized. By clustering transportation networks, we aim to classify them based on shared characteristics. This classification will provide valuable guidance for future researchers in selecting appropriate transportation networks for testing their methodologies. To address the complexity and interrelation of multiple features associated with transportation networks, we employ techniques from unsupervised learning for more effective representation of network groups.

The main contributions of this study are as follows: 
\begin{itemize}
	\item The incorporation of critical topological features of transportation networks, primarily derived from graph theory, focusing on network structure.
	
	\item The utilization of two different dimensionality reduction techniques to manage the large number of topological features and potential correlations among them.
	
	\item The adoption of two distinct unsupervised learning clustering methods to classify transportation networks, with a comparison of results using various clustering metrics.
	
	\item  A detailed discussion of the topological features and their influence on the resulting classification.
\end{itemize}

\section{Background}
\subsection{Network Science}
Network science is a field that focuses on modeling and analyzing various networks~\citep{muller1995neural, barabasi2013network}, spanning social networks~\citep{mitchell1974social}, molecular networks~\citep{bray2003molecular}, communication networks~\citep{monge2003theories}, and road networks~\citep{godfrey1969mechanism}. Urban road networks represent the spatial and geometrical relationships of roads and streets within cities~\citep{strano2013urban}. The study of urban networks has gained significant attention in recent years, thanks to advancements in geographic information systems (GIS), the increased availability of data from digital devices and sensors, and the development of more efficient and robust computational systems~\citep{badhrudeen2022geometric}. Given the growing utilization of transportation networks in recent studies, the classification of these networks has become a crucial task to aid scientists in selecting the most suitable networks for their research. In this context, network topology serves as a valuable source of information for this classification.

Network topology pertains to the arrangement of elements (nodes and links) within a network~\citep{casali2019topological}. The topological structure of networks typically encompasses essential information describing the network, including the number of links connected to nodes, the available paths between arbitrary nodes, and the significance of different nodes in the overall resilience of the network~\citep{wang2020road}. Various measures have been employed in network analysis studies to gain insights into the topology of networks~\citep{spadon2018topological}.

Fundamental topological characteristics of networks are related to the number of nodes and links in the network, and the length of links~\citep{porta2006network}. Additionally, another important group of characteristics includes centrality indices, which indicate the criticality of nodes within the network~\citep{koschutzki2005centrality}. Several centrality indices have been proposed in graph theory and widely used in network studies, such as degree centrality~\citep{merchan2020quantifying}, closeness centrality~\citep{shang2020robustness}, betweenness centrality~\citep{lin2017comparative}, and PageRank centrality~\citep{page1998pagerank}. Research has shown strong correlations between these centrality indices and network attributes, including disaster resilience, traffic propagation, and the efficiency of mobility. For instance, the betweenness centrality of a node measures the prevalence of shortest paths in the network that pass through the node between arbitrary nodes. Higher values of betweenness centrality indicate that the node plays a crucial role within the network~\citep{barthelemy2004betweenness}. Leveraging these centrality features can assist in categorizing networks into groups with similar characteristics. However, due to the challenges in interpreting the associated values with these network indices, coupled with the diversity of these indices, manual network classification is not feasible. A comprehensive approach with the capability to perform this task automatically with self-supervision is imperative.

\subsection{Unsupervised Learning}

In the past two decades, machine learning (ML) techniques have demonstrated their effectiveness across various domains of knowledge. In the field of transportation, ML methods have proven to be superior to conventional approaches in tasks such as traffic forecasting~\citep{li2018brief, schimbinschi2015traffic}, travel demand prediction~\citep{chu2018travel, koushik2020machine}, and autonomous vehicle navigation~\citep{sabzekar2023deep, mehditabrizi2023deep}. ML is generally divided into three sub-areas: supervised learning, unsupervised learning, and reinforcement learning. While supervised learning deals with labeled data, unsupervised learning operates with unlabeled data, aiming to derive insights from this raw data. In the transportation domain, unsupervised learning methods have exhibited their effectiveness in clustering various forms of public transportation big data~\citep{galba2013public}, GPS trajectories~\citep{reyes2020gps}, electric vehicle charging stations~\citep{straka2019clustering}, and demand patterns~\citep{liu2019exploring}.

\section{Problem Description}
The transportation network is represented as a directed graph, denoted as $ G (V, E, A)$, where $V= \{v_1, v_2, \dots , v_N \}$ is the set of nodes (i.e., intersections), $|V| = N$, $E$ is the set of links (i.e., roads connecting intersections) defined in Equation~\ref{EQ_E}, and $A$ is the adjacency matrix of the graph, defined in Equation~\ref{EQ_A}:
\begin{equation}\label{EQ_E}
	E = \{ (v_i, v_j) \mid \text{road connecting intersection } v_i \text{ to intersection } v_j \}
\end{equation}
\begin{equation}\label{EQ_A}
	 A_{ij} = 
	\begin{cases} 
		1 & \text{if } (v_i, v_j) \in E \\
		0 & \text{otherwise}
	\end{cases}
\end{equation}

The problem of transportation networks classification involves identifying a set of clusters, denoted as $ C = \{c_1, c_2, \dots, c_m\}$, where $m$ is the number of clusters, and each cluster $i$ ($i \in \{1, 2, \dots, m\}$), is defined as a set of networks, $c_i = \{G_1, G_2, \dots, G_r\}$, where $G_j$ belongs to cluster $i$. As the number of items in each cluster may vary, the value of $r$ is not consistent across different clusters. Clustering techniques aim to assign each network, $G_j$, to the most appropriate cluster, $c_i$, based on its similarities with other networks in $c_i$. We define a feature matrix, $F \in R^{k \times s}$, where $k$ is the number of networks to be clustered, and $s$ is the number of features per network. These features are computed through topological analysis of networks and are explained in detail in the following section. 

\section{Methodology}
\subsection{Network characteristics}
Networks are represented as graph-structured data, and graph theory encompasses a range of problems related to graphs. Within this framework, various characteristics have been defined to represent different properties of graphs. This paper focuses on topological characteristics, which are explained in the following sections, including general features and centrality indices.

\subsubsection{General Features}

\textbf{Number of nodes/links:}
Measure the total number of nodes and links in the transportation network. This provides a basic understanding of the network's scale.

\textbf{Average clustering coefficient (ACC):}
Determines the clustering coefficient for the network. This measures the extent to which nodes tend to cluster together. High clustering coefficients suggest the presence of tightly interconnected subgroups within the network. The average clustering coefficient for the graph $G$ is determined by

\begin{equation}\label{EQ_Average_Clustering_Coef}
	C = \frac{1}{n}\sum_{v \in G} c_v
\end{equation}

where $n$ is the number of nodes in $G$.

\textbf{Average shortest path length (ASPL):}
Calculates the ASPL between nodes in the network. It indicates the average number of steps required to travel between any two nodes. Shorter average path lengths imply better accessibility and connectivity within the transportation network. The average shortest path length is determined by 

\begin{equation}\label{EQ_Average_Shortest_Path_Length}
		a =\sum_{\substack{s,t \in V \\ s\neq t}} \frac{d(s, t)}{n(n-1)}
\end{equation}

where $V$ is the set of nodes in $G$, $d(s, t)$ is the shortest path from $s$ to $t$, and $n$ is the number of nodes in $G$.

\textbf{Diameter (longest shortest path):}
Determines the diameter of the network, which is the longest shortest path between any pair of nodes. It indicates the maximum distance one would need to travel to reach any other node in the transportation network. In order to calculate diameter, we recall that diameter is the maximum eccentricity, and the eccentricity of a node $v$ is the maximum distance from $v$ to all other nodes in graph $G$. 

\textbf{Radius (shortest shortest path):}
Network radius is a concept that measures the extent of reach or influence from a central point within a network. The network radius indicates the shortest distance from a central node to the farthest node within the network. In transportation terms, we can visualize the network radius as the distance a traveler starting from a central hub would need to cover to reach the outermost point in the network. It gives us an idea of how widely the influence or connectivity of that central hub extends within the network. In another definition, the radius is the minimum eccentricity. 

\textbf{Density:}
Calculates the network density, which measures the proportion of actual connections to the total number of possible connections in the network. Dense networks indicate a high level of connectivity. Network density is determined by
\begin{equation}\label{EQ_Network_Density}
	d = \frac{m}{n(n-1)}
\end{equation}
where $n$ is the number of nodes, and $m$ is the number of edges in Network $G$.

\textbf{Number of weakly and strongly connected components (WCCs / SCCs):}
Counts the number of weakly and strongly connected components. In a directed graph, a weakly connected component is a subset of nodes wherethere is a path between every pair of nodes in the subset, while ignoring the directions of the edges. In a directed graph, a strongly connected component is a subset of nodes where there is a directed path between every pair of nodes in the subset. 

\textbf{Size of the giant weakly and strongly connected components (GWCC / GSCC):}
Determines the number of nodes in largest weakly and strongly connected components. 

\textbf{Average global efficiency (AGE):}
Determines the average global efficiency of the network. The efficiency of a pair of nodes in a network is the multiplicative inverse of the shortest path distance between the nodes. The average global efficiency of a network is the average efficiency of all pairs of nodes. Global efficiency measures how efficiently information or resources flow across the entire network

\textbf{Average local efficiency (ALE):}
Local efficiency focuses on the efficiency of communication within local neighborhoods or clusters. The local efficiency of a node in the network is the average global efficiency of the sub-network induced by the neighbors of the node. The average local efficiency is the average of the local efficiencies of each node.

\textbf{Reciprocity:}
The ratio of mutual connections (connections where both nodes interact with each other) to the total number of connections in the network. The reciprocity coefficient can range from $0$ to $1$. Network reciprocity is determined by
\begin{equation}\label{EQ_Network_Reciprocity}
	r = \frac{|{(u,v) \in G|(v,u) \in G}|}{|{(u,v) \in G}|}
\end{equation}

\textbf{Transitivity:}
Network transitivity quantifies the tendency for triangles to form within the network. In simpler terms, network transitivity examines how interconnected a node's immediate connections are with each other. Transitivity measures the ratio of closed triangles to all connected triples in a network. A closed triangle consists of three nodes that are interconnected in a triangular formation, meaning each node is connected to the other two. Network transitivity is defined by
\begin{equation}\label{EQ_Network_Transitivity}
	T = 3 \frac{\# triangles}{\# triads}
\end{equation}
\textbf{Degree assortativity coefficient (DAC)}: 
Assortativity measures the similarity of connections in the graph with respect to the node degree. In other words, assortativity measures the tendency of nodes to connect to others with similar or dissimilar characteristics. In the context of transportation networks, assortativity analysis can reveal patterns of connections based on attributes such as geographical location, transportation mode, or capacity.

\subsubsection{Centrality Indices}

\textbf{In-degree centrality (IC):}
The in-degree centrality for a node $v$ is the fraction of nodes its incoming edges are connected to. 

\textbf{Out-degree centrality (OC):}
The out-degree centrality for a node $v$ is the fraction of nodes its outgoing edges are connected to.

\textbf{Closeness centrality (CC):}
Closeness centrality of a node $u$ is the reciprocal of the average shortest path distance to $u$ over all $n-1$ reachable nodes. Closeness centrality is defined by
\begin{equation}\label{EQ_Closeness_Centrality}
	CC(u) = \frac{n - 1}{\sum_{v=1}^{n-1} d(v, u)},
\end{equation}
where $d(v, u)$ is the shortest-path distance between $v$ and $u$, and $n-1$ is the number of nodes reachable from $u$. 

\textbf{Betweenness centrality (BC):}
Betweenness centrality of a node $v$ is the sum of the fraction of all-pairs shortest paths that pass through $v$ and is defined by
\begin{equation}\label{EQ_Betweenness_Centrality}
	BC(v) =\sum_{s,t \in V} \frac{\sigma(s, t|v)}{\sigma(s, t)}
\end{equation}
where $V$ is the set of nodes, $\sigma(s, t)$ is the number of shortest (s, t)-paths, and $\sigma(s, t|v)$ is the number of those paths passing through some node $v$ other than $s, t$. If $s=t$, then $\sigma(s, t) = 1$, and if $v \in {s, t}$, then $\sigma(s, t|v) = 0$.

\textbf{Eigenvector centrality (EC):}
Eigenvector centrality is a measure used to determine the importance of a node in a network. In the context of a transportation network, eigenvector centrality helps identify the most critical transportation hubs. It considers not only the number of connections a node has but also the importance of the nodes to which it is connected. Mathematically, the eigenvector centrality $EC_i$ of a node $i$ in a transportation network can be calculated by
\begin{equation}\label{EQ_EigenVector_Centrality}
	EC_i = \frac{1}{\lambda} \sum_{j} A_{ij}x_j
\end{equation}
where $EC_i$ represents the eigenvector centrality of node $i$, $\lambda$ is the dominant eigenvalue of the adjacency matrix, and $A_{ij}$ is the element of the adjacency matrix that represents the connection between node $i$ and node $j$. The eigenvector centrality helps identify critical nodes that are not only well-connected but are also connected to other important nodes, making them vital for the overall transportation flow in the network.

\textbf{PageRank centrality (PC):}
PageRank is a method used to evaluate the importance of nodes based on the concept that important nodes are likely to be linked to by other important nodes. It helps in identifying critical transportation hubs that are pivotal for the efficient flow of traffic within the network. The formula for PageRank in a transportation domain can be represented as follows:
\begin{equation}\label{EQ_PageRank}
	PC(p_i) = \frac{1-d}{N} + d \sum_{p_j \in M(p_i)} \frac{PC(p_j)}{L(p_j)}
\end{equation}
where $PC(p_i)$ represents the PageRank of node $p_i$, $N$ is the total number of nodes in the network, $d$ is the damping factor that represents the probability of following a link, $M(p_i)$ is the set of nodes that link to $p_i$, and $L(p_j)$ is the number of outbound links from node $p_j$.

\subsection{Scaling}
Features calculated based on the network characteristics and centrality indices exhibit varying scales. Continuing with this scenario, the accuracy of downstream tasks, such as dimensionality reduction and clustering, may be adversely affected~\citep{ahsan2021effect}. To mitigate potential inaccuracies, we employ a scaling method. In this context, we opt for min-max scaling with a target range of $(0, 1)$ for all features, as calculated by the following equations:

\begin{equation}\label{}
	X_{\text{std}} = 
		\frac{X - X.\text{min}}{X.\text{max} - X.\text{min}}
\end{equation}

\begin{equation}\label{EQ_PageRank}
	X_{\text{scaled}} = X_{\text{std}} \times (\text{max} - \text{min}) + \text{min}
\end{equation}

where $X.\text{min}$ and $X.\text{max}$ represent the minimum and maximum values in the original range of the data, respectively. Similarly, $\text{min}$ and $\text{max}$ indicate the minimum and maximum values in the scaled range. Finally, $X_{\text{scaled}}$ denote the scaled value of the feature. This scaling method linearly transforms the features into a fixed range, ensuring that the largest occurring data point corresponds to the maximum value and the smallest one corresponds to the minimum value.

\subsection{Dimensionality Reduction}
The evolution of data collection systems has granted access to larger datasets, often featuring numerous features per data record. While these features may contain valuable information, some may exhibit high correlation, leading to potential redundancy~\citep{fan2014challenges}. Additionally, the abundance of features incurs high computational costs which is an undesirable aspect. Furthermore, in clustering tasks, obtaining a visual representation of data points in a two-dimensional space proves beneficial for determining an optimal cluster count and interpreting clustering results~\citep{davidson2002visualizing}. To address these considerations, we employ two distinct dimensionality reduction methods: Principal Component Analysis (PCA)~\citep{pearson1901liii} and Isometric Feature Mapping (ISOMAP)~\citep{tenenbaum2000global}, aiming to project our features into a two-dimensional feature set.

\subsubsection{Principal Component Analysis (PCA)}
PCA stands as one of the most widely used linear dimension reduction algorithms~\citep{jolliffe2016principal}. It operates through a projection-based approach, transforming data by projecting it onto a set of orthogonal axes. The fundamental premise of PCA lies in maximizing the variance or spread of data in the lower-dimensional space while mapping data from a higher-dimensional space. The principal components, constituting linear combinations of the original variables, are determined by eigenvectors satisfying the principle of least squares~\citep{abdi2010principal}.

PCA is computed through the following steps:

\begin{enumerate}
	\item Compute the covariance matrix of the data matrix.
	
	\item Calculate the eigenvalues and eigenvectors of the covariance matrix.
	
	\item Sort the eigenvalues in decreasing order.
	
	\item Select the top $n$ principal components, capturing the most variance in the data.
	
	\item Project the data onto the selected principal components.
	
\end{enumerate}

While PCA is effective, interpreting principal components becomes challenging when dealing with a large number of variables. It is most suitable when variables exhibit a linear relationship, and susceptibility to significant outliers should be noted.

\subsubsection{Isometric Feature Mapping (ISOMAP)}
In contrast to PCA, ISOMAP represents a non-linear dimensionality reduction method. It offers a straightforward technique for estimating the intrinsic geometry of a data manifold and embedding the data in a lower-dimensional space. The algorithm derives an estimate of each data point's neighbors on the manifold, facilitating an efficient and widely applicable approach to various data sources and dimensionalities~\citep{van2009dimensionality}.

ISOMAP is computed through the following steps:

\begin{enumerate}
	\item Build a neighborhood graph: Identify the k-nearest neighbors for each data point and create edges between points that are mutual k-nearest neighbors.
	
	\item Compute geodesic distances between all data points: Geodesic distance, the shortest path on the neighborhood graph between two points, is calculated.
	
	\item Compute the embedding matrix: Utilize classical multidimensional scaling (MDS) to compute an embedding matrix, representing each data point in a lower-dimensional space.
	
\end{enumerate}

\subsection{Clustering Methods}
In this paper, we employ two distinct clustering approaches: K-means and Hierarchical Density-Based Spatial Clustering of Applications with Noise (HDBSCAN).

\subsubsection{K-means}
The K-means algorithm clusters data by attempting to group samples into $n$ clusters with equal variance, minimizing a criterion known as inertia or within-cluster sum-of-squares. This algorithm requires specifying the number of clusters and performs effectively on large sample sizes, finding application across various fields.
The K-means algorithm partitions a set of $N$ samples, $X$, into $K$ disjoint clusters, $C$, each characterized by the mean, $\mu_i$, of the samples in the cluster. These means are commonly referred to as cluster "centroids"; note that they are not necessarily points from $X$, although they exist in the same space. Minimization of inertia is defined by
\begin{equation}
	\sum_{i=0}^{n}\min_{\mu_j \in C}(||x_i - \mu_j||^2)
\end{equation}

\subsubsection{Hierarchical Density-Based Spatial Clustering of Applications with Noise (HDBSCAN)}
HDBSCAN performs DBSCAN over varying epsilon values and integrates the result to identify a clustering that provides the best stability over epsilon. This enables HDBSCAN to identify clusters of varying densities, unlike DBSCAN, making it more robust to parameter selection. DBSCAN views clusters as regions of high density separated by areas of low density, allowing clusters found by DBSCAN to take any shape, in contrast to K-means, which assumes clusters are convex. The central concept in DBSCAN is core samples, located in areas of high density. A cluster is a set of core samples, each close to one another (measured by some distance), and a set of non-core samples that are close to a core sample but are not core samples themselves. The algorithm has two parameters, min samples and eps, which formally define what is considered dense. Higher min samples or lower eps indicate higher density required to form a cluster.

\subsection{Evaluation Metrics}\label{Sec_Evaluation_Metrics}
In this section, we examine three widely used evaluation metrics for clustering tasks.

\textbf{Silhouette Coefficient}

A higher Silhouette Coefficient score indicates a model with better-defined clusters. The Silhouette Coefficient is computed for each sample and comprises two scores~\citep{rousseeuw1987silhouettes}:

$a$: The mean distance between a sample and all other points in the same class. \\
$b$: The mean distance between a sample and all other points in the next nearest cluster.

The Silhouette Coefficient \( s \) for a single sample is then given by:

\begin{equation}
	s = \frac{b - a}{\max(a, b)}
\end{equation}

The Silhouette Coefficient for a set of samples is the mean of the Silhouette Coefficient for each sample. \\

\textbf{Calinski and Harabasz Score}

Also known as the Variance Ratio Criterion, the score is defined as the ratio of the sum of between-cluster dispersion to within-cluster dispersion. For a dataset $E$ of size $n_E$ clustered into $K$ clusters, the Calinski-Harabasz score $s$ is defined as the ratio of the mean between-clusters dispersion and within-cluster dispersion~\citep{calinski1974dendrite}:

\begin{equation}
	s = \frac{\mathrm{tr}(B_k)}{\mathrm{tr}(W_k)} \times \frac{n_E - k}{k - 1}
\end{equation}

where $\mathrm{tr}(B_k)$ is the trace of the between-group dispersion matrix, and $\mathrm{tr}(W_k)$ is the trace of the within-cluster dispersion matrix defined by:
\begin{equation}
	W_k = \sum_{q=1}^k \sum_{x \in C_q} (x - c_q) (x - c_q)^T
\end{equation}
\begin{equation}
	B_k = \sum_{q=1}^k n_q (c_q - c_E) (c_q - c_E)^T
\end{equation}

with $C_q$ being the set of points in cluster $q$, $c_q$ the center of cluster $q$, $c_E$ the center of $E$, and $n_q$ the number of points in cluster $q$. \\

\textbf{Davies-Bouldin Index}

A lower Davies-Bouldin index indicates a model with better separation between clusters. This index represents the average similarity between clusters, where similarity is a measure comparing the distance between clusters with the size of the clusters. Zero is the lowest possible score, and values closer to zero indicate a better partition~\citep{davies1979cluster}.

The index is defined as the average similarity between each cluster $C_i$ for $i=1,\dots,k$ and its most similar one $C_j$. In the context of this index, similarity is defined as a measure $R_{ij}$ that balances:

$s_i$: The average distance between each point of cluster $i$ and the centroid of that cluster (also known as cluster diameter).

$d_{ij}$: The distance between cluster centroids $i$ and $j$.

A simple choice to construct $R_{ij}$ so that it is non-negative and symmetric is:

\begin{equation}
	R_{ij} = \frac{s_i + s_j}{d_{ij}}
\end{equation}

Then the Davies-Bouldin index is defined as:
\begin{equation}
	DB = \frac{1}{k} \sum_{i=1}^k \max_{i \neq j} R_{ij}
\end{equation}

\subsection{Data}

\begin{figure}
	\centering
	\includegraphics[width=170mm]{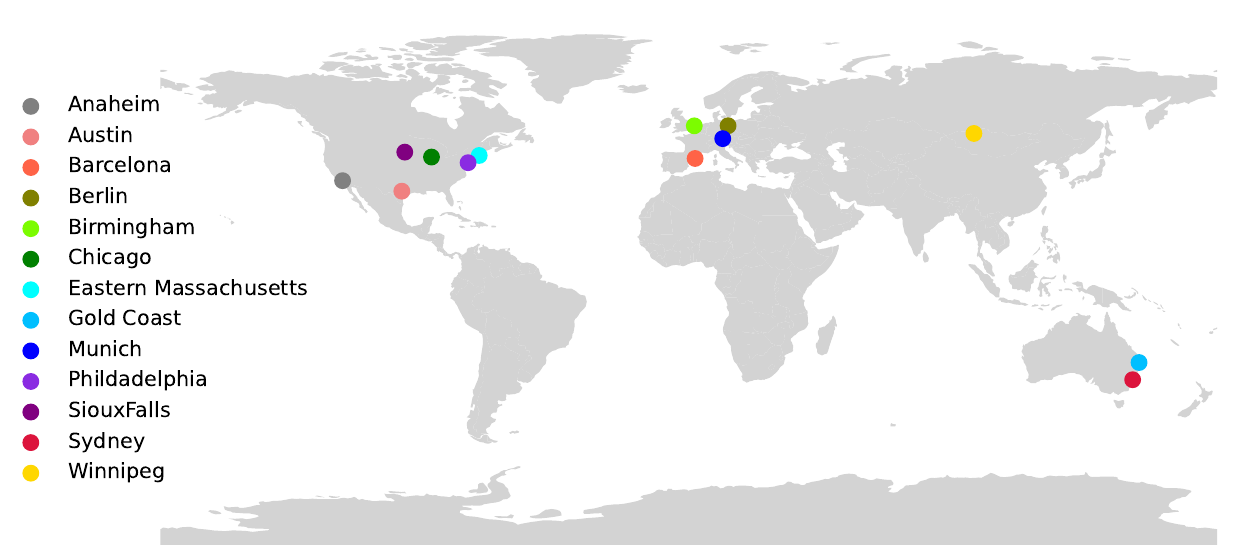}
	\caption{Transportation networks projected on a world map.}
	\label{Fig_Map}
\end{figure}

The dataset of transportation networks is provided by the Transportation Networks for Research Core Team~\citep{Dataset_Transportation}. This collection includes transportation networks for 14 renowned cities worldwide (\autoref{Fig_Map}). The datasets consist of network structures (node-link relations), node locations, origin-destination trip data, and, for some of them, traffic assignment data. These networks include Anaheim~\citep{Dataset_Anaheim}, Austin~\citep{Dataset_Austin}, Barcelona~\citep{Dataset_Transportation}, Berlin~\citep{Dataset_Berlin}, Birmingham~\citep{Dataset_Birmingham}, Chicago Sketch and Regional~\citep{Dataset_Chicago_1, Dataset_Chicago_2}, Eastern Massachusetts~\citep{Dataset_EasternMassachusetts}, Golden Coast~\citep{Dataset_GoldenCoast}, Philadelphia~\citep{Dataset_Philadelphia}, SiouxFalls~\citep{Dataset_SiouxFalls}, Sydney~\citep{Dataset_Sydney}, and Winnipeg~\citep{Dataset_Winnipeg}. Chicago-Sketch is a fairly realistic yet aggregated representation of the Chicago-Region network.

\section{Results and Discussion}

\subsection{Network Characteristics}

\begin{figure}
	\centering
	\includegraphics[width=100mm]{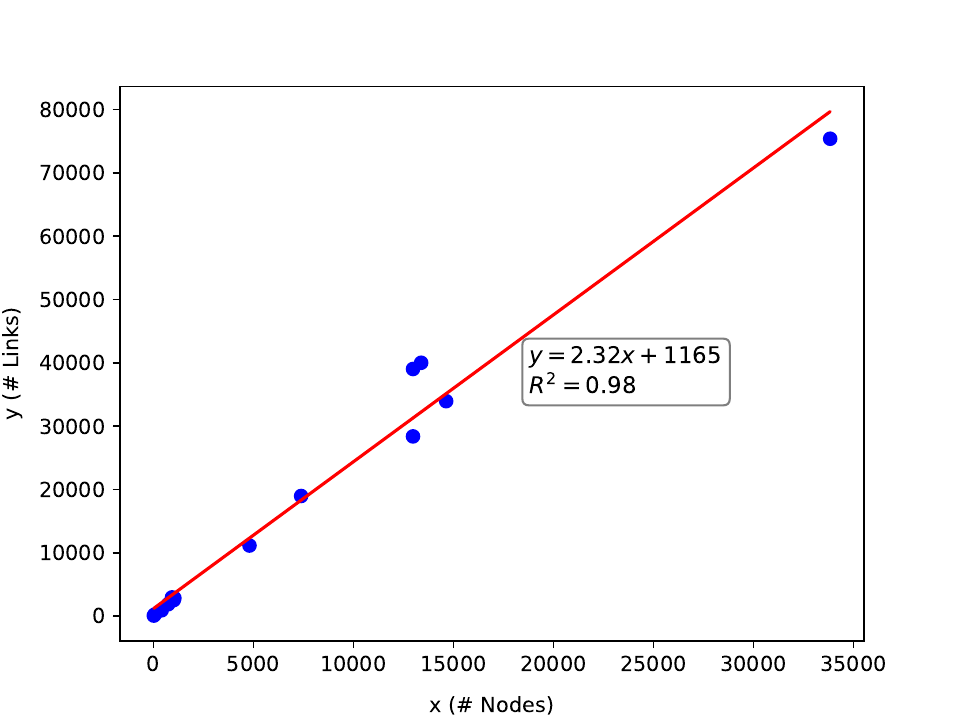}
	\caption{Relationship between number of nodes and number of links in the transportation networks.}
	\label{Fig_Node_Link}
\end{figure}

\autoref{Table_Network_Characteristics_1} and \autoref{Table_Network_Characteristics_2} present the general network characteristics for the test networks used in this study. In \autoref{Table_Network_Characteristics_3}, we present centrality indices, each of which denotes the average values of centrality indices across all nodes in the network. Additionally, we established a linear regression model to estimate the number of links in the network based on the number of nodes. \autoref{Fig_Node_Link} illustrates the resulting line, expressed by $y = 2.32 x + 1165$, where $x$ and $y$ are the number of nodes and links, respectively. This line, with an $R^2$ of $0.98$, demonstrates a high level of accuracy. The coefficient of $2.32$ in the formula indicates that, on average, each node is connected to $2.32$ links. In a similar study~\citep{badhrudeen2022geometric}, the formula for the fitted line was $y = 1.33 x + 2907$, suggesting that, on average, each node is connected to $1.33$ links. The distinction lies in the fact that the authors in that study considered urban road networks, which typically include all classifications of roads from freeways and highways to secondary roads and even alleys. In contrast, the transportation networks used in this study specifically represent primary roads in urban regions, excluding other road types.

\begin{table}
	\centering
	\small
	\caption{Network Characteristics (General Features)}
	\label{Table_Network_Characteristics_1}
	\renewcommand{\arraystretch}{1.4}
	\begin{tabular}{lllllllll}
		\hline
		Network & Nodes & Links & Link Length & Density & Diameter & Radius & Reciprocity & Transitivity \\
		\hline
		Anaheim & 416 & 914 & 0.510 & 0.005 & 31 & 16 & 0.613 & 0.035 \\
		Austin & 7388 & 18961 & 0.593 & 0 & 118 & 73 & 0.883 & 0.014 \\
		Barcelona & 1020 & 2522 & 0.645 & 0.002 & 31 & 20 & 0.574 & 0.069 \\
		Berlin & 12981 & 28376 & 0.052 & 0 & 116 & 72 & 0.486 & 0.216 \\
		Birmingham & 14639 & 33937 & 1.205 & 0 & 139 & 69 & 0.765 & 0.089 \\
		Chicago-Regional & 12982 & 39018 & 0.693 & 0 & 111 & 58 & 0.943 & 0.046 \\
		Chicago-Sketch & 933 & 2950 & 2.778 & 0.003 & 32 & 18 & 1 & 0.066 \\
		Eastern-Massachusetts & 74 & 258 & 8.55 & 0.048 & 9 & 5 & 1 & 0.223 \\
		Gold Coast & 4807 & 11140 & 0.243 & 0 & 135 & 70 & 0.931 & 0.043 \\
		Munich & 742 & 1872 & 0.502 & 0.003 & 56 & 29 & 1 & 0.028 \\
		Philadelphia & 13389 & 40003 & 0.457 & 0 & 98 & 59 & 0.938 & 0.013 \\
		SiouxFalls & 24 & 76 & 4.132 & 0.138 & 6 & 4 & 0.966 & 0.052 \\
		Sydney & 33837 & 75379 & 0.246 & 0 & 24 & 13 & 0.578 & 0.006 \\
		Winnipeg & 1052 & 2836 & 0.748 & 0.003 & 40 & 29 & 0.578 & 0.138 \\
		\hline
	\end{tabular}
\end{table}

\begin{table}
	\centering
	\small
	\caption{Network Characteristics (General Features-Continued)}
	\label{Table_Network_Characteristics_2}
	\renewcommand{\arraystretch}{1.4}
	\begin{tabular}{llllllllll}
		\hline
		Network & GSCC & GWCC & SCCs & WCCs & ACC & DAC & ASPL & AGE & ALE \\
		\hline
		Anaheim & 416 & 416 & 1 & 1 & 0.049 & 0.28 & 11.868 & 0.11 & 0.065 \\
		Austin & 7381 & 7388 & 8 & 1 & 0.01 & 0.245 & 48.627 & 0.028 & 0.011 \\
		Barcelona & 929 & 930 & 92 & 91 & 0.058 & -0.005 & 14.009 & 0.077 & 0.062 \\
		Berlin & 12842 & 12981 & 140 & 1 & 0.152 & 0.169 & 50.649 & 0.024 & 0.158 \\
		Birmingham & 14560 & 14578 & 41 & 28 & 0.088 & 0.258 & 44.243 & 0.027 & 0.093 \\
		Chicago-Regional & 12978 & 12979 & 5 & 4 & 0.037 & -0.013 & 44.711 & 0.029 & 0.037 \\
		Chicago-Sketch & 933 & 933 & 1 & 1 & 0.034 & -0.061 & 12.676 & 0.105 & 0.037 \\
		Eastern-Massachusetts & 74 & 74 & 1 & 1 & 0.287 & -0.169 & 4.465 & 0.288 & 0.308 \\
		Gold Coast & 4783 & 4783 & 25 & 25 & 0.031 & 0.247 & 54.081 & 0.027 & 0.031 \\
		Munich & 742 & 742 & 1 & 1 & 0.017 & 0.209 & 19.716 & 0.073 & 0.017 \\
		Philadelphia & 13389 & 13389 & 1 & 1 & 0.012 & 0.162 & 43.173 & 0.029 & 0.013 \\
		SiouxFalls & 24 & 24 & 12 & 12 & 0.052 & 0.162 & 3.011 & 0.427 & 0.053 \\
		Sydney & 32956 & 32956 & 14 & 14 & 0.005 & 0.216 & 81.574 & 0.016 & 0.005 \\
		Winnipeg & 1040 & 1040 & 1057 & 109 & 0.013 & -0.028 & 18.848 & 0.069 & 0.085 \\
		\hline
	\end{tabular}
\end{table}

\begin{table}
	\centering
	\caption{Network Characteristics (Centrality Indices)}
	\label{Table_Network_Characteristics_3}
	\renewcommand{\arraystretch}{1.4}
	\small
	\begin{tabular}{lllllll}
		\hline
		Network & IC & OC & CC & BC & EC & PC \\ 
		\hline
		Anaheim & 0.00530 & 0.00530 & 0.08680 & 0.03090 & 0.01970 & 0.00240 \\
		Austin & 0.00030 & 0.00030 & 0.02090 & 0.00670 & 0.00130 & 0.00010 \\
		Barcelona & 0.00240 & 0.00240 & 0.06550 & 0.01220 & 0.00600 & 0.00100 \\
		Berlin & 0.00017 & 0.00017 & 0.02092 & 0.00394 & 0.00033 & 0.00008 \\
		Birmingham & 0.00020 & 0.00020 & 0.02680 & 0.00310 & 0.00050 & 0.00010 \\
		Chicago-Regional & 0.00023 & 0.00023 & 0.02284 & 0.00352 & 0.00075 & 0.00008 \\
		Chicago-Sketch & 0.00339 & 0.00339 & 0.08003 & 0.01466 & 0.01162 & 0.00107 \\
		Eastern-Massachusetts & 0.04776 & 0.04776 & 0.22951 & 0.07386 & 0.08752 & 0.01351 \\
		GoldCoast & 0.00048 & 0.00048 & 0.01873 & 0.01134 & 0.00097 & 0.00021 \\
		Munich & 0.00340 & 0.00340 & 0.05288 & 0.02792 & 0.01164 & 0.00135 \\
		Philadelphia & 0.00022 & 0.00022 & 0.02347 & 0.00330 & 0.00078 & 0.00007 \\
		SiouxFalls & 0.13768 & 0.13768 & 0.33630 & 0.16712 & 0.18337 & 0.04167 \\
		Sydney & 0.00007 & 0.00007 & 0.01411 & 0.00247 & 0.00015 & 0.00003 \\
		Winnipeg & 0.00257 & 0.00257 & 0.05280 & 0.01844 & 0.00307 & 0.00095 \\
		\hline
	\end{tabular}
\end{table}

\subsection{Dimensionality Reduction}
\begin{figure}
	\centering
	\includegraphics[width=100mm]{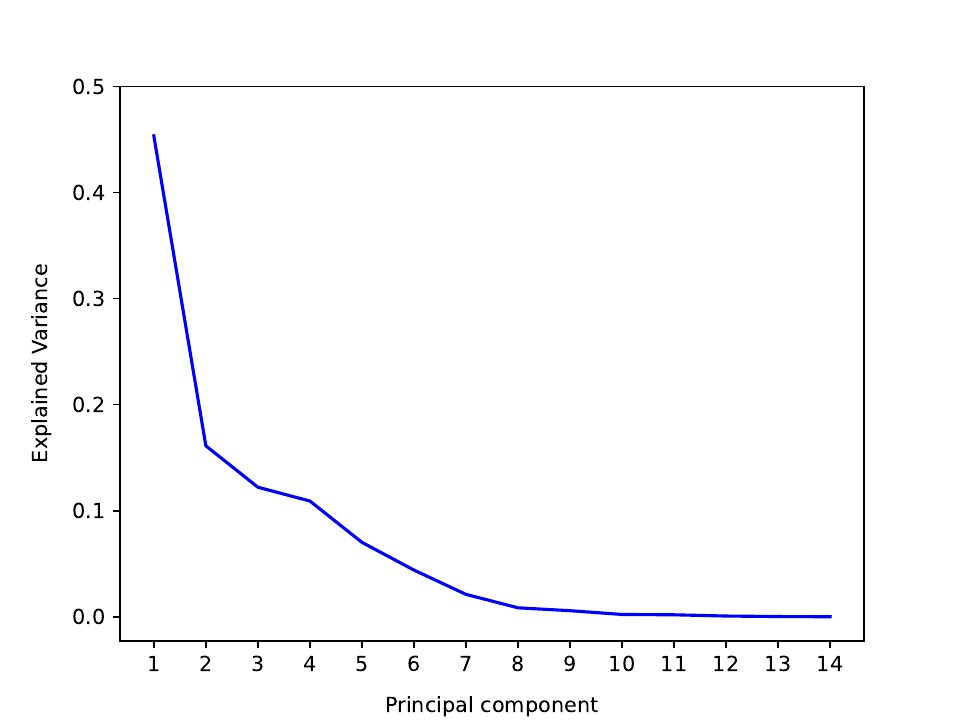}
	\caption{Explained variance of each principal component in PCA method.}
	\label{Fig_Principal_Component}
\end{figure}

This section presents the results of adopting dimensionality reduction methods. \autoref{Fig_Principal_Component} displays the outcome of implementing PCA. The principal components are ordered based on the variance they explain, with the first principal component accounting for the most variance. The explained variance graph of PCA illustrates the percentage of the total variance explained by each principal component. In this study, the first two principal components explain most of the variance, suggesting that using only these two components may be sufficient for subsequent analysis.

\begin{figure}
	\centering
	\includegraphics[width=170mm]{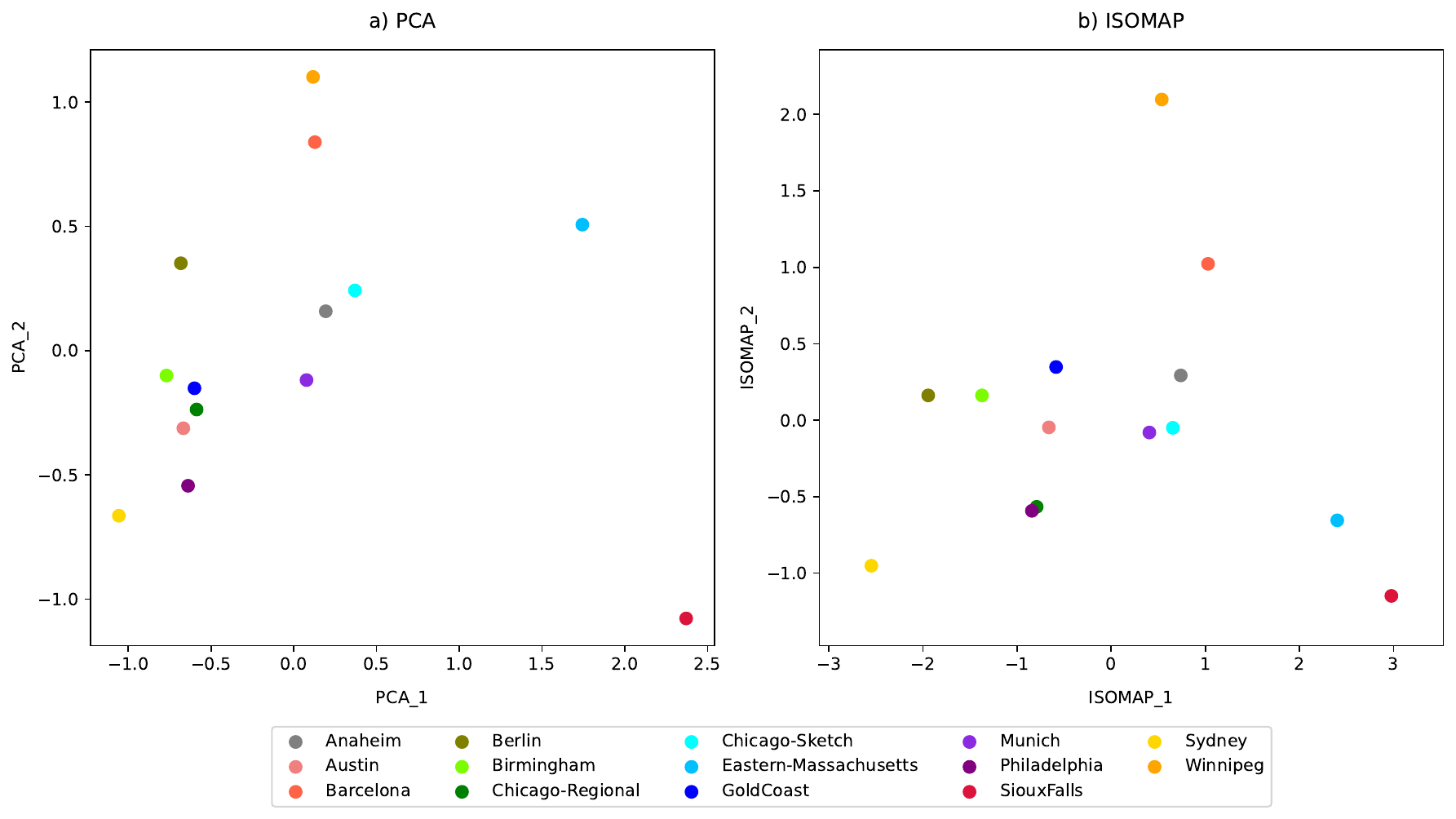}
	\caption{Projected space of dimensionality reduction methods: a) PCA, and b) ISOMAP}
	\label{Fig_PCA_ISOMAP}
\end{figure}

\autoref{Fig_PCA_ISOMAP} illustrates the outcomes of the dimensionality reduction methods, PCA, and ISOMAP, as applied to transportation networks characteristics. Examining the projected space generated by PCA and ISOMAP, it is clear that clusters are more intuitively recognizable in the results obtained from PCA than in those from ISOMAP. While this intuitive assessment is useful, our determination of which dimensionality reduction method is superior must rely on numerical analysis, thoroughly explored in the following subsection.

\subsection{Clustering Performance}
This section evaluates the clustering methods and the upstream dimensionality reduction approaches. Given that two dimensionality reduction approaches and two clustering methods are utilized, four combinations are evaluated using the metrics described in \autoref{Sec_Evaluation_Metrics}, with the results presented in \autoref{Table_Metrics}. Larger values in Silhouette and Calinski-Harabasz scores indicate better performance, while the Davies-Bouldin score indicates the best models with the lowest values. Based on the results in this Table, the combination of PCA followed by the K-means clustering method outperforms the other three combinations. Consequently, this combination is selected for further analysis of results.

\begin{figure}
	\centering
	\includegraphics[width=100mm]{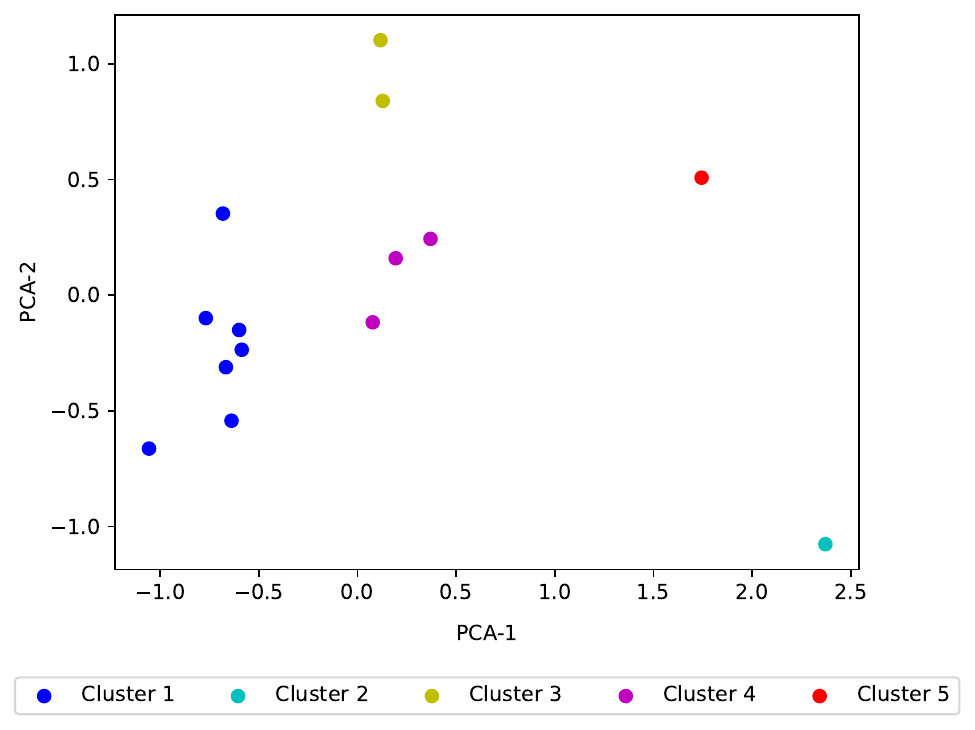}
	\caption{Visualization of clustering outcome of using method K-means / PCA.}
	\label{Fig_Kmeans_Result}
\end{figure}

The associated results of the K-means / PCA clustering are depicted in \autoref{Fig_Kmeans_Result}. The figure shows that five clusters are chosen, and data points in each cluster are closely grouped. Cluster 2 and Cluster 5 include only one data point, indicating that these data points are unique and cannot be placed in other clusters.

\autoref{Table_Clusters} lists the transportation networks classified in each cluster obtained from K-means / PCA. Cluster 1 comprises the transportation networks of large urban areas, characterized by a large number of nodes and links. These networks exhibit higher resilience when facing congestions or disruptions in critical nodes, as they offer more alternative links and paths. Cluster 2 exclusively accommodates the SiouxFalls dataset, a publicly known and widely used transportation network for various purposes. While SiouxFalls is valuable, it represents a rough approximation of the city, making it less suitable for real-world applications. Barcelona and Winnipeg are classified in Cluster 3, representing medium-sized transportation networks with virtually identical features. Cluster 4 encompasses small-sized transportation networks, including Anaheim, Chicago-Sketch, and Munich. Chicago-Sketch is widely adopted in traffic assignment studies as a test network. These networks in Cluster 4 are suitable choices as test networks, offering affordable computational costs while preserving real-world characteristics. Lastly, Cluster 5 is solely occupied by the Eastern-Massachusetts network, similar to SiouxFalls in its wide usage as a test network. Eastern-Massachusetts is slightly larger than SiouxFalls and differs in appearance, leading our clustering method to categorize them differently.

\begin{table}[h]
	\centering
	\small
	\renewcommand{\arraystretch}{1.4}
	\caption{Clustering evaluation metrics.}
	\label{Table_Metrics}
	\begin{tabular}{llll}
		\hline
		Method & Silhouette score & Calinski Harabasz score & Davies Bouldin score \\
		\hline
		K-means / PCA & \textbf{0.510} & \textbf{37.674} & \textbf{0.297} \\
		K-means / ISOMAP & 0.461 & 27.655 & 0.579 \\
		HDBSCAN / PCA & 0.393 & 8.747 & 1.066 \\
		HDBSCAN / ISOMAP & 0.155 & 0.931 & 9.211 \\
		\hline
	\end{tabular}
\end{table}

\begin{table}[h]
	\centering
	\small
	\caption{Classification of transportation networks.}
	\label{Table_Clusters}
	\renewcommand{\arraystretch}{1.4}
	\begin{tabular}{ll}
		\hline
		Cluster & Networks \\
		\hline
		Cluster 1 & Austin, Berlin, Birmingham, Chicago-Regional, Gold Coast, Philadelphia, Sydney \\
		Cluster 2 & SiouxFalls \\
		Cluster 3 & Barcelona, Winnipeg \\
		Cluster 4 & Anaheim, Chicago-Sketch, Munich \\
		Cluster 5 & Eastern-Massachusetts \\
		\hline
	\end{tabular}
\end{table}

\begin{figure}
	\centering
	\includegraphics[width=170mm]{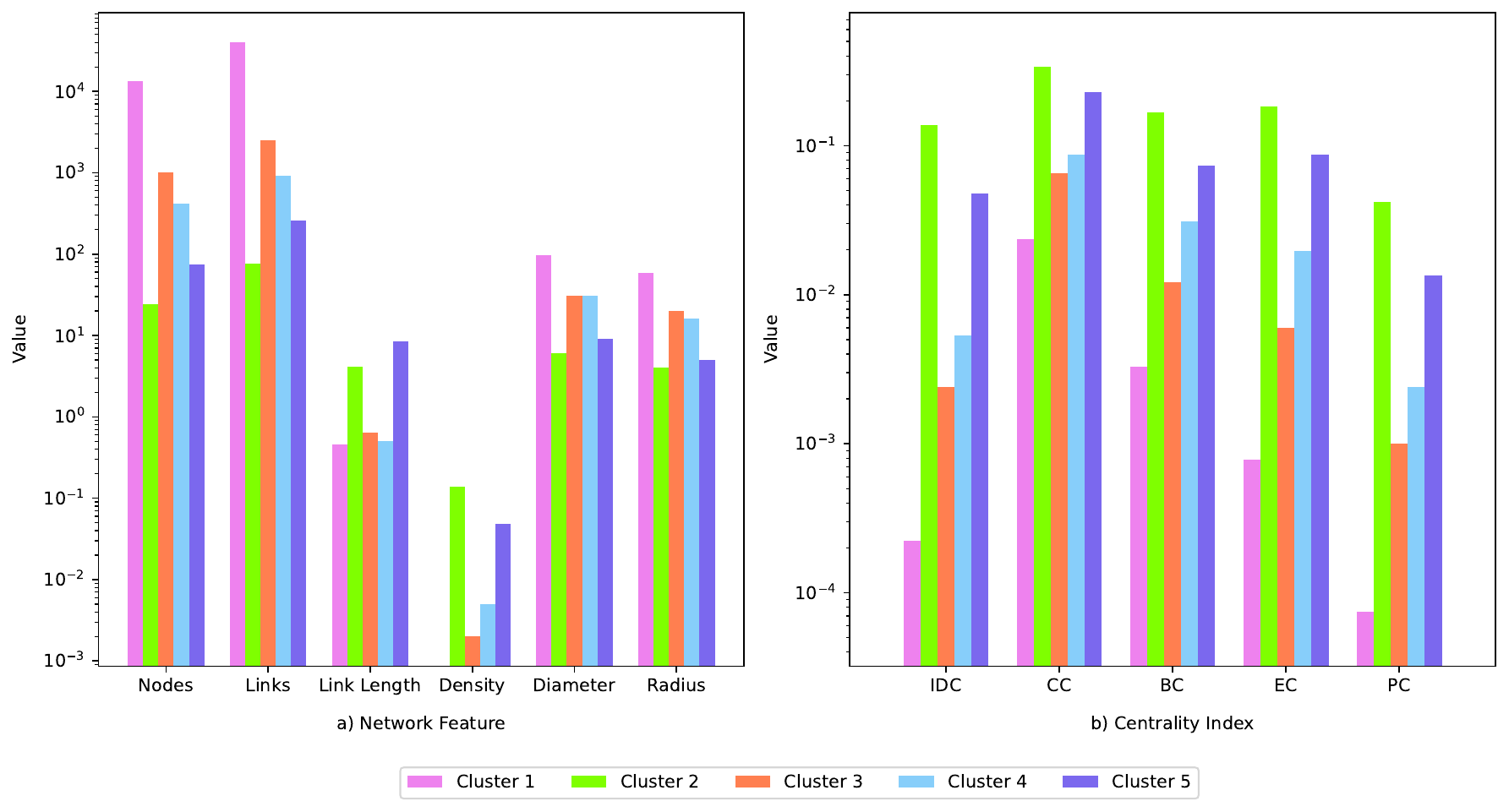}
	\caption{Network characteristics in each cluster of transportation networks: a) Network Feature, and b) Centrality Index}
	\label{Fig_Features_Clusters}
\end{figure}

\autoref{Fig_Features_Clusters} illustrates the average values of network characteristics for each cluster. Sub-figures a) and b) depict network features and centrality indices, respectively. As shown in the figure, features, including the number of nodes and links, diameter, and radius, follow the same pattern, indicating that Cluster 1 and Cluster 2 have the highest and lowest values, respectively, with other clusters ordered similarly. For features such as link length and density, Cluster 1 shows the lowest values. However, the highest values in these two features do not follow the same pattern. While Cluster 5 shows the highest value of link length, Cluster 2 has the highest value of density. Based on mathematical formulations, diameter and radius are dependent on the number of links. The results of clustering also highlight this fact, indicating that Cluster 1, with the highest number of links, has the highest values for diameter and radius. Moreover, diameter and radius are much similar in definition, which is shown in the figure.

Furthermore, the results of centrality indices, shown in \autoref{Fig_Features_Clusters}-b, validate the outcomes of clustering. If our clustering method didn’t work well, we would expect to see values of a specific centrality index to be similar among two different colors. However, this discrepancy is not observed in the results. Similar to \autoref{Fig_Features_Clusters}-a, \autoref{Fig_Features_Clusters}-b follows the pattern that centrality indices have their lowest values for networks in Cluster 1 and the highest values for Cluster 2. Centrality values of cluster 1 are low compared to other clusters, indicating that there are fewer nodes that are placed on the shortest paths between two other nodes. In other words, nodes in the networks of cluster 1 are less critical compared to nodes in other clusters, since the resiliency of the networks in cluster 1 is higher, meaning that there are several alternative paths to move from one node to another. On the other hand, centrality indices for cluster 2 (SiouxFalls) have their highest values, indicating that some nodes in this network are very critical and play a key role in the total resiliency of the network. If one of those nodes becomes congested or fails, the overall performance of the network will be severely affected. The key observation in both of these sub-figures in \autoref{Fig_Features_Clusters} is that values of each network characteristic differ among clusters, confirming the effectiveness of the clustering method used.

The obtained results of this study can assist authors of future studies. Our Cluster 1 of transportation networks is composed of large-scale networks with high resiliency characteristics among nodes. Studies that focus on adopting high computational centrality indices, such as betweenness and closeness centrality, are suggested to use transportation networks of Clusters 3 or 4 to mitigate the unbearable long run-times. Also, if a study aims to assess the relationship between a centrality measure and a characteristic of the network, such as traffic congestion, and investigates this relationship for different transportation networks, the authors of the study should choose their test networks from different clusters obtained in this study. In other words, choosing two networks such as Barcelona and Winnipeg and claiming that a proposed method is effective on different transportation networks is not a valid claim, as both of these networks are classified in identical cluster (Cluster 3) and have similar characteristics.

A general recommendation is that authors of future studies investigate the effectiveness of their proposed approaches on at least three transportation networks: one on SiouxFalls or Eastern-Massachusetts, which are very helpful for debugging the methods due to their low computational demands; one on a medium-scale network from Cluster 3 or 4, and finally, one on a network from Cluster 1 to demonstrate the effectiveness of their proposed method on a network close to the real world.

\section{Conclusion and Future work}
In this study, we present a comprehensive framework for the classification of transportation networks based on their topological features. We calculate various network characteristics derived from the topological structure of the network, encompassing both network features and centrality indices. Subsequently, we employ two dimensionality reduction methods, PCA and ISOMAP. These methods contribute to reducing dimensions while retaining the most valuable features and combining highly correlated ones.

Following dimensionality reduction, we apply two clustering approaches, namely K-means and HDBSCAN, to achieve an unsupervised classification of fourteen transportation networks. Various metrics are employed to assess the accuracy of clustering methods. K-means / PCA outperforms the other three combinations of dimensionality reduction and clustering methods, achieving a Silhouette metric score of $0.510$. This method results in the identification of $5$ clusters within the networks. An analysis of the topological characteristics of the networks in each cluster validates the effectiveness of this classification for the fourteen networks.

Future works may involve expanding the number of transportation networks for classification, incorporating additional network features, and exploring alternative clustering approaches that may yield improved results.

\bibliographystyle{unsrtnat}
\bibliography{references}

\end{document}